# Subset Labeled LDA for Large-Scale Multi-Label Classification

Yannis Papanikolaou*
Grigorios Tsoumakas†

**Abstract**

Labeled Latent Dirichlet Allocation (LLDA) is an extension of the standard unsupervised Latent Dirichlet Allocation (LDA) algorithm, to address multi-label learning tasks. Previous work has shown it to perform in par with other state-of-the-art multi-label methods. Nonetheless, with increasing label sets sizes LLDA encounters scalability issues. In this work, we introduce Subset LLDA, a simple variant of the standard LLDA algorithm, that not only can effectively scale up to problems with hundreds of thousands of labels but also improves over the LLDA state-of-the-art. We conduct extensive experiments on eight data sets, with label sets sizes ranging from hundreds to hundreds of thousands, comparing our proposed algorithm with the previously proposed LLDA algorithms (Prior–LDA, Dep–LDA), as well as the state of the art in extreme multi-label classification. The results show a steady advantage of our method over the other LLDA algorithms and competitive results compared to the extreme multi-label classification algorithms.

## 1 Introduction

Multi-label learning addresses supervised learning problems where each instance can be tagged with more than one labels at the same time [20]. Apart from tasks that fall inherently into this category, such as image or text classification, prior work [24, 1] has shown that recommendation and ranking problems can also be formulated as multi-label learning problems. A great body of prior work has dealt with developing methods for multi-label tasks [29, 22, 17, 16, 23, 9, 10], however the majority of these algorithms struggle to scale to problems having more than a few thousand labels. Since the ever increasing flow and volumes of data in modern-day applications directly affect the area of multi-label learning, algorithms that can effectively and efficiently scale up to large-scale problems are required.

Extreme multi-label classification is an emerging field that attempts to address the above challenge, by proposing algorithms that can tackle problems with extremely large label sets ($> 10^4$ labels).

We modify an already existing algorithm, Labeled Latent Dirichlet Allocation (LLDA) to successfully deal with such tasks. LLDA was introduced by [16] as an extension of standard, unsupervised Latent Dirichlet Allocation (LDA) [4], to deal with multi-label learning tasks. LLDA proceeds exactly as its unsupervised predecessor with the only difference that during training LLDA constraints the topics for each instance to be the instance's labels. [18] have proposed two extensions of LLDA, the first incorporating label frequencies in the corpus (Prior–LDA) and the second taking into account label dependencies (Dep–LDA). In Section 2.2, we describe the above algorithms in detail.

Apart from delivering results competitive with state-of-the-art multi-label algorithms, LLDA's training is by design particularly fit for large-scale and extreme learning problems. Specifically, unlike most other algorithms, during training LLDA's time complexity is not dependent of the label set size $L$, but on the label cardinality $\overline{L_M}$, i.e. the average number of labels assigned per instance. Given that $\overline{L_M}$ is typically in the order of $10^1$, it is valid to assume that LLDA's training complexity will only be affected by the training set size and the number of features per instance. On the other hand, during testing, LLDA is equivalent to LDA and the algorithm is linearly dependent of $L$, which makes the algorithm unfit for large-scale multi-label classification.

In order to make LLDA appropriate for tasks with very large $L$, we propose an extension to LLDA, Subset LLDA. We conduct extensive experiments on four small scale data sets and four large scale data sets, with $L$ ranging from 101 to 670,000 comparing our approach to Prior–LDA and Dep–LDA as well as two of the top performing extreme classification algorithms, FastXML and PfastreXML. The results show a consistent advantage of our method compared to the other LLDA algorithms and competitive results with the extreme classification methods. Our motivation by introducing Subset LLDA, is to contribute one more method to the extreme classification inventory, that may be more apt than other methods for specific experimental scenarios.

The rest of the paper is organized as follows: In Section 2 we present the existing methods in the literature to address extreme classification tasks and the LDA and LLDA algorithms. Section 3 introduces our proposed algorithm, Subset LLDA. In Section 4 we describe the experiments and the results and in Section 5 we conclude with the implications of our results and possible future directions.

---
*Aristotle University of Thessaloniki
†Aristotle University of Thessaloniki

## 2 Background and Related Work

In this section, we first describe the main methods in the literature to tackle extreme multi-label learning problems and then present the theoretical background regarding our proposed method and more specifically LDA, LLDA, and the other two LLDA extensions, Prior–LDA and Dep–LDA. Note that throughout the paper we assume that the Collapsed Gibbs Sampling (CGS) algorithm [6] is employed for all LLDA methods.

**2.1 Extreme Classification Methods** Algorithms aiming to tackle extreme classification tasks take one of the following approaches:

- learning a hierarchy of the label set and solve the training and prediction procedures locally at each node. In this case, we start by having a root node containing the entire label set and then partition each node's label set to its respective child, according to some optimization criterion, or more broadly some partitioning formulation. The nodes are recursively partitioned until each leaf contains only a small number of labels. During prediction, a testing instance is passed down the tree until it reaches one or more leaf nodes. In this way, a given multi-label classifier will be trained and predict only on a small subset of the label set leading to sub-linear or even logarithmic (if the hierarchy is balanced) complexities.

- construct an embedding of the output space in a lower dimension. Embedding-based methods render training and prediction tractable by assuming that the training label matrix is low-rank, reducing the label set size by projecting the high dimensional label vectors onto a low dimensional linear subspace.

Hierarchy of Multi-label classifiers (HOMER) [22] has been the first method to follow the label partitioning approach, constructing the label hierarchy by performing a recursive clustering of the label set. Label Partitioning by Sublinear Ranking (LPSR) [24] proceeded by learning an input partition and a label assignment to each partition of the space, optimizing precision at k. [15] presented FastXML, which learns an ensemble of trees. Nodes are split by learning a separating hyperplane which partitions training points in two sub-categories. FastXML optimizes the normalized Discounted Cumulative Gain (nDCG) such that each training point's relevant labels are ranked as highly as possible in its partition. PfastreXML [7] is an extension of FastXML that replaces the nDCG loss function with its propensity scored variant.

In the embedding-based methods, LEML [27] formulates the problem as that of learning a low-rank linear model, by studying the multi-label problem in the empirical risk minimization (ERM) framework. SLEEC [3], proceeds by learning a small ensemble of local distance preserving embeddings, focusing on improving prediction on rare labels.

Recently two more methods were proposed that do not fall in any of the above categories. DiSMEC [2] is a large-scale distributed framework for learning one-versus-rest linear classifiers, using a double layer of parallelization. PD-sparse [26] uses the assumption that for each instance there are only a few correct labels and show that a simple margin-maximizing loss yields an extremely sparse dual solution. The authors propose a fully-corrective block coordinate Frank-Wolfe algorithm to solve the primal-dual sparse penalties.

**2.2 LDA and LLDA** Let us denote as $L$ the number of labels, $l$ being a label and $V$ the number of features, $v$ being a feature type and $w_i$ being a feature token at position $i$ of the instance. $M$ is the number of instances ($M_{TRAIN}$ and $M_{TEST}$ will represent the training and testing set sizes respectively), $m$ being an instance. Also, $L_m$ will denote the number of $m$'s labels and $N_m$ the number of its non-zero features. The notation is summarized in Table 1.

LDA assumes that, given a set of instances, there exist two sets of distributions, the label-features distributions named $\boldsymbol{\phi}$ and the instances-labels distributions named $\boldsymbol{\theta}$[1]. CGS LDA marginalizes out $\boldsymbol{\phi}$ and $\boldsymbol{\theta}$, and uses only on the latent variable assignments $\mathbf{z}$. The algorithm employs two count matrices during sampling, the number of times that $v$ is assigned to $l$ across the data set, represented by $n_{lv}$ and the number of feature tokens in $m$ that have been assigned to $l$, represented by $n_{ml}$. During sampling, CGS updates the hard assignment $z_i$ of $w_i$ to one of $l \in \{1...L\}$. This update is performed sequentially for all tokens in the data set, for a fixed number of iterations. The update equation giving the probability of setting $z_i$ to label $l$, conditional on $w_i$, $m$, the hyperparameters $\boldsymbol{\alpha}$ and $\boldsymbol{\beta}$, and the current label assignments of all other word tokens (represented by $\cdot$) is:

---

[1] Specifically, LDA is defining $\boldsymbol{\phi}$ and $\boldsymbol{\theta}$ in terms of topics since it is unsupervised, but to ease understanding we consider through the paper that topics and labels are equivalent.

| | |
|---:|:---|
| $V$ | the size of the vocabulary |
| $L$ | the number of labels |
| $M_{Train}$ | the number of training instances |
| $M_{Test}$ | the number of instances to be predicted |
| $m$ | a instance |
| $v$ | a feature type |
| $w_i$ | a single feature token, i.e. an instance of a feature type |
| $l$ | a label |
| $z_i$ | the label assignment to a token $w_i$ |
| $N_d$ | number of tokens in $m$ |
| $\mathcal{L}_m$ | set of labels in $m$ |
| $n_{lv}$ | number of times that type $v$ has been assigned to $l$ across the data set |
| $n_{ml}$ | number of tokens in $m$ that have been assigned to $l$ |
| $\phi_l$ | feature type distribution for $l$ |
| $\theta_m$ | label distribution for instance $m$ |
| $\alpha_l$ | Dirichlet prior hyperparameter on $\theta$ for $l$ |
| $\beta$ | Dirichlet prior hyperparameter on $\boldsymbol{\phi}$ |

Table 1: Notation used throughout the article.

$$(2.1) \quad p(z_i = l \mid w_i = v, m, \boldsymbol{\alpha}, \boldsymbol{\beta}, \cdot) \propto \frac{n_{lv \neg i} + \beta}{\sum_{v'=1}^{V} (n_{lv' \neg i} + \beta_{v'})} \cdot \frac{n_{ml \neg i} + \alpha_l}{N_m + \sum_{l'=1}^{L} \alpha_{l'}}.$$

During training, LLDA constrains the possible assignments for a token to a label to the instance's observed labels. Therefore, during training, the sampling update of CGS is:

$$(2.2) \quad p(z_i = l \mid w_i = v, m, \boldsymbol{\alpha}, \boldsymbol{\beta}, \cdot) \propto \begin{cases} \frac{n_{lv \neg i} + \beta}{\sum_{v'=1}^{V} (n_{lv' \neg i} + \beta_{v'})} \cdot \frac{n_{ml \neg i} + \alpha_l}{N_m + \sum_{l'=1}^{L} \alpha_{l'}}, & \text{if } l \in \mathcal{L}_m \\ 0, & \text{otherwise.} \end{cases}$$

Inference on test instances is performed similarly to unsupervised CGS - LDA the label–features distributions, $\boldsymbol{\phi}$, being fixed to those previously learned and estimated on the training data, and then the test instances' $\boldsymbol{\theta}$ distributions are estimated.

To estimate the $\boldsymbol{\phi}$ and $\boldsymbol{\theta}$ parameters, we employ the $CGS^p$ equations, presented by [14], that employ the full distribution over feature tokens. These methods calculate the expected values of the standard CGS estimators and are therefore expecially suited for the large-scale setting that we are addressing in this work, since they allow us to achieve better performance by drawing fewer samples than with the standard estimators. For the calculation of the estimators we refer the interested reader to the pseudocode of the relevant paper.

LLDA employs a symmetric $\alpha$ hyperparameter over labels, giving equal weight on them. Nevertheless, in most real-world tasks labels tend to have skewed distributions. Moreover, modeling label dependencies can improve performance [17], especially as the number of labels increases. To address these issues, [18] have proposed Prior–LDA and Dep–LDA respectively. Prior–LDA incorporates the label frequencies observed in the training set via an asymmetric $\alpha$ hyperparameter: a frequent label will have a larger $\alpha_l$ value than a rare one. Specifically, it is set to

$$(2.3) \quad \alpha_l = \eta \cdot f_l + \alpha$$

with $\eta, \alpha$ being user defined parameters and $f_l$ representing the frequency of $l$ in the training corpus, $f_l \in [0, 1]$.

Dep–LDA is a two-stage algorithm: first, an unsupervised LDA model is trained with $T$ labels and using as training data, each instance's label set[2]. The estimated LDA model will incorporate information about the label dependencies, since relevant labels will tend to be described by the same topic(s). Second, an LLDA model is trained. During prediction, the previously estimated $\boldsymbol{\theta}', \boldsymbol{\phi}'$ parameters of the unsupervised LDA model are used to calculate an asymmetrical $\alpha_{dk}$. Specifically, the $\boldsymbol{\alpha_d}$ vector will be

$$(2.4) \quad \boldsymbol{\alpha_d} = \eta(\boldsymbol{\theta}'_{\boldsymbol{d}} \cdot \boldsymbol{\phi}') + \alpha$$

## 3 Subset LLDA

When dealing with tasks with very large $L$ ($> 10^4$), LLDA and its extensions can not scale in a satisfying manner since they are linearly dependent on $L$ during prediction. To alleviate this, we propose a simple extension to standard LLDA during prediction, by constraining the label space in which the algorithm can search for solutions. Specifically, our method proceeds in two stages:

- First, for each test instance, a set of candidate labels is determined. A number of approaches can be followed to determine this candidate list, for simplicity we retrieve the $n$-most relevant instances from the training set and set the candidate list as the union of the retrieved instances tags.

- Second, we predict with LLDA, but constrain the possible labels to the previously determined candidate labels list.

---
[2]i.e., the feature tokens of each instance are its labels

Constraining the label set during prediction can also be useful for an additional reason. In that phase, LLDA needs to search the entire label space to recommend labels for a given test instance. Since this is a probabilistic method, the algorithm may converge to local optima, especially as $L$ is increasing. To concretize, let us consider a trained LLDA model for which a specific feature $v$ has a high probability $\phi_{lv}$ for several labels. Also, let us consider a test instance $m$ that contains $v$, for which only one of the aforementioned labels is semantically relevant. In that case, it is possible that these noisy labels, coupled with LLDA's probabilistic nature, will lead the algorithm to favor one of the irrelevant labels at the expense of the correct label. This problem can of course be relieved by averaging over many samples and Markov Chains (MC) [18, 14], but in most real cases this is too expensive time-wise.

To constrain the label set of a given instance $m$, we choose a very simple approach, always with the aim to provide the minimum time overhead to our algorithm. We retrieve the $n$-most relevant training instances to $m$ and use the union of their respective label sets as the candidate list. Formally, $L_m = \bigcup_{i=1}^{n} L_{m_i}$, with $m \in M_{TEST}, m_i \in M_{TRAIN}$. To retrieve the most relevant training instances we use the tf-idf representation for each instance and employ the cosine similarity [11] keeping the ten most relevant training instances for each test instance.

**3.1 Time complexity** The CGS algorithm for LDA proceeds as follows: for every word token of every instance in the corpus, it calculates probability distribution over all labels and then samples a label for the token, out of this calculated probability. In this way, stadard LDA is linearly dependent on $L$. Formally, it will be

$$(3.5) \qquad T_{LDA} \propto \mathcal{O}(M \cdot \overline{V_m} \cdot L).$$

LLDA, introduces supervision during training, by constraining the possible labels that a word token can get on the instance's label set $L_d$. Formally, during training LLDA's complexity will be

$$(3.6) \qquad T_{LLDA} \propto \mathcal{O}(M_{TRAIN} \cdot \overline{V_m} \cdot \overline{L_m}).$$

As explained, during testing LLDA is equivalent to LDA so its complexity will be given by Equation 3.5. To alleviate this, with our approach we constrain Subset LLDA during testing, to only consider labels from the $\nu$-most relevant instances. The total complexity of Subset LLDA, involves also finding the $\nu$-most relevant instances which in our cases will be $\mathcal{O}(M_{TEST} \cdot M_{TRAIN})$:

$$(3.7)$$
$$T_{SubsetLLDA} \propto \mathcal{O}(M_{TEST} \cdot \overline{V_m} \cdot \nu \cdot \overline{L_m} + M_{TEST} \cdot M_{TRAIN}).$$

## 4 Empirical Evaluation

We here present the experiments that we carried out by presenting the data sets, the evaluation measures, the experimental setup and the results. We compare Subset LLDA with its LLDA counterparts, Prior–LDA and Dep–LDA, as well as with two of the best performing extreme multi-label classification algorithms, Fast XML and PfastreXML. Our results for the large-scale tasks are directly comparable to the ones provided in the extreme classification repository[3], therefore it is also possible to compare our method with the rest of the extreme classification methods proposed in the literature (e.g. SLEEC, LEML, LPSR, DiSMEC, PD-SPARSE).

**4.1 Data Sets** In our experiments, we employed four small scale and four large scale data sets, their statistics being illustrated in Table 2. $Bibtex$ [8], $Delicious$ [22], $Mediamill$ [19] and $EUR-Lex$ [12] were retrieved from the Mulan library website[4]. We used the train/test split provided in the web site and have removed from each data set all instances having empty label sets. $Wiki10$, $Delicious - 200k$ and $Amazon - 670k$ were retrieved from the extreme classification repository [5]. Finally, the $BioASQ$ data set is a subset of the data set provided in the BioASQ challenge[6].The exact data set used in the experiments is available upon request to the authors and will be also made available in the extreme classification web page.

---
[3]http://manikvarma.org/downloads/XC/XMLRepository.html
[4]http://mulan.sourceforge.net/datasets-mlc.html
[5]http://manikvarma.org/downloads/XC/XMLRepository.html
[6]http://participants-area.bioasq.org/general_information/Task5a/

|  | instances | | Labels | | | |
| --- | --- | --- | --- | --- | --- | --- |
| Data set | $D_{Train}$ | $D_{Test}$ | $K$ | $\overline{|\mathcal{K}_d|}$ | Avg. Freq. | $V$ |
| Bibtex | 4,880 | 2,515 | 159 | 2.38 | 73.05 | 1,836 |
| Delicious | 12,910 | 3,181 | 983 | 19.06 | 250.34 | 500 |
| Mediamill | 30,993 | 12,914 | 101 | 4.38 | 1902.12 | 120 |
| EUR-Lex | 15,539 | 3,809 | 3,993 | 5.31 | 25.73 | 5,000 |
| Wiki10 | 14,146 | 6,616 | 30,938 | 8.52 | 18.64 | 101,938 |
| BioASQ | 40,000 | 10,000 | 19,218 | 13.05 | 174.97 | 36,480 |
| Delicious-200k | 196,606 | 100,095 | 205,443 | 72,29 | 75,54 | 782,585 |
| Amazon-670k | 490,449 | 153,025 | 670,091 | 3.99 | 5.45 | 135,909 |

Table 2: Statistics for the data sets used in the experiments. Column "Avg. Freq." refers to the average label frequency. All figures concerning labels and word types are given for the respective training sets.

**4.2 Evaluation** We consider four performance measures: the micro-averaged F1 measure (Micro-F), the macro-averaged F1 measure (Macro-F)[21], precision at k (precision@k) and propensity scored precision at k (PS precision@k) . In the following, we describe them briefly.

**Macro-F and Micro-F scores** These two measures represent a weighted function of precision and recall, and emphasize the need for a model to perform well in terms of both of these underlying measures. The Macro-F score is the average of the F1-scores that are achieved across all labels, while the Micro-F score is the average F1 score weighted by each label's frequency. Macro-F tends to emphasize performance on infrequent labels, while Micro-F tends to emphasize performance on frequent labels.

**Precision@k and Propensity Scored Precision@k** Precision@k counts the number of correct predictions in the top k positive predictions. It is calculated as

$$(4.8) \qquad precision@k = \frac{1}{k} \sum_{l \in rank_k \hat{y}} y_l$$

while PS precision@k is given by:

$$(4.9) \qquad PSprecision@k = \frac{1}{k} \sum_{l \in rank_k \hat{y}} \frac{y_l}{p_l}$$

with $\hat{y}$ representing the predicted label set vector, $rank_k \hat{y}$ returning the k largest indices of $y$ ranked in descending order and $p_l$ being the propensity score for label $l$ [7].

**4.3 Setup** We here describe the setup for the algorithms used throughout the experiments. Fast XML and PfastreXML were used with default parameters, with the relevant software packages provided in the extreme classification repository[7].

Regarding the LLDA models, we provide our Java implementation[8]. Since the differences of the algorithms lie in the way they perform prediction on new data, we trained the same model for all algorithms in order to ensure fairness of comparison, keeping symmetrical the $\alpha_l$ parameter with $\alpha_l = \frac{50}{K}$. For Dep–LDA, we additionally need to train an LDA model to calculate the hyperparameter on $\theta$ (ref. to Equation 2.4). For its training we use 100 topics and 200 iterations, 50 burn-in iterations and we set $\alpha = 0.1$, $\beta = 0.01$.

During prediction, for Prior–LDA we set $\eta = 50$, $\alpha = \frac{30}{L}$ and for Dep–LDA, we set $\eta = 120$, $\alpha = \frac{30}{L}$ across all data sets[9]. In both training and prediction and across all data sets, we used one MC, 200 iterations, a burn-in period of 50 iterations and a lag of 5 iterations between each sample. All samples were averaged to obtain the respective parameter estimates for each method. The $\beta$ parameter was set to 0.01.

Since all algorithms output rankings of relevant labels for each instance, in order to obtain a hard assignment of labels to instances to compute the Micro-F and Macro-F scores, we used the *rcut* thresholding strategy introduced by [25], which sets as a threshold the cardinality of the training set. Finally, all experiments were run on an 16-core Intel Xeon CPU, with 2.27GHz and 36Gb of RAM.

**4.4 Results** In Tables 3–8 we report the results of our experiments. For the LLDA methods, we report the average over five runs.

---
[7]http://manikvarma.org/code/FastXML/download.html and http://manikvarma.org/code/PfastreXML/download.html
[8]https://www.dropbox.com/s/rypmlt18zk6jxdh/sllda.zip?dl=0
[9]Initial experiments did not reveal significant differences with different parameterizations.

|  | FastXML | PfastreXML | PriorLDA | DepLDA | Subset LLDA |
|---|---|---|---|---|---|
| Bibtex | 0.382 | **0.397** | 0.363 | 0.314 | 0.384 |
| Delicious | **0.347**▽ | 0.322 | 0.255 | 0.273 | 0.304 |
| Mediamill | 0.577 | 0.572 | 0.511 | 0.512 | **0.580** |
| EUR-Lex | 0.413 | 0.436 | 0.321 | 0.357 | **0.443**▽ |
| BioASQ | 0.382 | 0.370 | - | - | **0.428**▽ |
| Wiki10 | 0.317 | **0.33**▽ | - | - | 0.283 |
| Delicious-200k | **0.162**▽ | 0.129 | - | - | 0.135 |
| Amazon | 0.192 | **0.316**▽ | - | - | 0.243 |
| Avg. Rank | 2.125 | 2.000 |  |  | 1.875 |

Table 3: Micro-F results for the LLDA-based and the extreme classification methods. A ▽ indicates a statistically significant difference between Subset LLDA and the best performing method, with a z-test and a significance level of 0.05. The - sign is used if the algorithm could not deliver predictions after 48 hours.

|  | FastXML | PfastreXML | PriorLDA | DepLDA | Subset LLDA |
|---|---|---|---|---|---|
| Bibtex | 0.273 | 0.288 | 0.257 | 0.204 | **0.292** |
| Delicious | 0.153 | **0.165**▽ | 0.100 | 0.090 | 0.136 |
| Mediamill | 0.070 | **0.101**▽ | 0.027 | 0.027 | 0.068 |
| EUR-Lex | 0.427 | 0.416 | 0.305 | 0.336 | **0.444** |
| BioASQ | 0.392 | 0.399 | - | - | **0.451**▽ |
| Wiki10 | **0.305**▽ | 0.285 | - | - | 0.272 |
| Delicious-200k | **0.105**▽ | 0.067 | - | - | 0.078 |
| Amazon | 0.426 | 0.483 | - | - | **0.486** |
| Avg. Rank | 2.125 | 2.000 |  |  | 1.875 |

Table 4: Macro-F results for the LLDA-based and the extreme classification methods. A ▽ indicates a statistically significant difference between Subset LLDA and the best performing method, with a z-test and a significance level of 0.05. The - sign is used if the algorithm could not deliver predictions after 48 hours.

|  | FastXML | PfastreXML | PriorLDA | DepLDA | Subset LLDA |
|---|---|---|---|---|---|
| Bibtex | **0.645**▽ | 0.565 | 0.551 | 0.501 | 0.579 |
| Delicious | **0.705**▽ | 0.546 | 0.488 | 0.526 | 0.525 |
| Mediamill | **0.873**▽ | 0.867 | 0.794 | 0.794 | 0.813 |
| EUR-Lex | 0.637 | 0.644 | 0.622 | 0.631 | **0.663**▽ |
| BioASQ | 0.176 | **0.189**▽ | - | - | 0.077 |
| Wiki10 | **0.825**▽ | 0.757 | - | - | 0.773 |
| Delicious-200k | **0.432**▽ | 0.261 | - | - | 0.163 |
| Amazon | 0.286 | **0.368**▽ | - | - | 0.350 |
| Avg. Rank | 1.625 | 2.000 |  |  | 2.375 |

Table 5: Precision@1 results for the LLDA-based and the extreme classification methods. A ▽ indicates a statistically significant difference between Subset LLDA and the best performing method, with a z-test and a significance level of 0.05. The - sign is used if the algorithm could not deliver predictions after 48 hours.

First, let us consider the differences among the LLDA methods. Subset LLDA is steadily outperforming both Prior–LDA and Dep–LDA in all settings and for all measures. It should be noted here, that Dep–LDA by design would benefit by averaging over more samples and more than one MC more than the other methods, since it employs the parameters learned from an unsupervised LDA model to calculate the hyperparameters for $\theta$, therefore it will be more prone than the other algorithms to get stuck in local optima. In other words, the LDA model introduces an additional factor of uncertainty, which could be alleviated with more samples and chains. We nevertheless restrict our experiments to only one MC and relatively few samples (thirty) since our main goal is to address large-scale tasks in which multiple MC averaging and averaging over many samples is not feasible. One more interesting observation, is that for tasks with few labels (*Bibtex*, *Mediamill*), Dep–LDA performs equally or worse to Prior LDA which may be explained by the fact that modeling dependencies does not necessarily help improving performance in small scale tasks.

|  | FastXML | PfastreXML | PriorLDA | DepLDA | Subset LLDA |
|---|---|---|---|---|---|
| Bibtex | **0.286**▽ | 0.254 | 0.238 | 0.215 | 0.243 |
| Delicious | **0.596**▽ | 0.482 | 0.391 | 0.416 | 0.427 |
| Mediamill | 0.531 | 0.534 | 0.480 | 0.471 | **0.539** |
| EUR-Lex | 0.425 | 0.445 | 0.332 | 0.358 | **0.457**▽ |
| BioASQ | 0.170 | 0.169 | - | - | **0.174** |
| Wiki10 | 0.570 | **0.572** | - | - | 0.560 |
| Delicious-200k | **0.362**▽ | 0.262 | - | - | 0.201 |
| Amazon | 0.195 | **0.320**▽ | - | - | 0.246 |
| Avg. Rank | 2.000 | 1.875 |  |  | 2.125 |

Table 6: Precision@5 results for the LLDA-based and the extreme classification methods. A ▽ indicates a statistically significant difference between Subset LLDA and the best performing method, with a z-test and a significance level of 0.05. The - sign is used if the algorithm could not deliver predictions after 48 hours.

|  | FastXML | PfastreXML | PriorLDA | DepLDA | Subset LLDA |
|---|---|---|---|---|---|
| Bibtex | **0.497** | 0.475 | 0.451 | 0.394 | 0.459 |
| Delicious | **0.330**▽ | 0.322 | 0.272 | 0.276 | 0.285 |
| Mediamill | **0.663**▽ | 0.661 | 0.600 | 0.600 | 0.615 |
| EUR-Lex | 0.239 | **0.396**▽ | 0.297 | 0.301 | 0.308 |
| BioASQ | 0.134 | **0.210**▽ | - | - | 0.070 |
| Wiki10 | 0.099 | **0.189**▽ | - | - | 0.122 |
| Delicious-200k | **0.065** | 0.031 | - | - | 0.039 |
| Amazon | 0.118 | **0.286**▽ | - | - | 0.241 |
| Avg. Rank | 1.875 | 1.625 |  |  | 2.500 |

Table 7: Propensity Scored Precision@1 results for the LLDA-based and the extreme classification methods. A ▽ indicates a statistically significant difference between Subset LLDA and the best performing method, with a z-test and a significance level of 0.05. The - sign is used if the algorithm could not deliver predictions after 48 hours.

Comparing Subset LLDA with the extreme classification methods, we observe that results are more mixed across data sets and evaluation measures. By considering the average rank per evaluation measure (last row of the Tables), we observe that PfastreXML has the upperhand in three out of six evaluation measures (precision@5, PS precision @1 and @5) and FastXML is ranked first for precision@1. Our algorithm, Subset LLDA, achieves the first place for Micro-F and Macro-F. Regarding the data set size, for small scale data sets Fast XML is ranked first in eleven out of 24 cases (four datasets and six evaluation measures), PfastreXML is ranked first in six cases and Subset LLDA is in the first place in seven cases. For the large scale data sets, Subset LLDA is ranked first in five out of 24 cases, PfastreXML is ranked first in eleven cases, while Fast XML in eight cases.

From these empirical results, we can make the following observations. First, the fact that PfastreXML is optimized for the propensity nDCG measure may explain why it outperforms the other two methods for the PS precision@k. Another important remark suggested by these results, is that Subset LLDA could be better suited for classification tasks while Fast XML and PfastreXML for ranking and retrieval tasks, since it performs better than the other two algorithms for Micro-F and Macro-F while being outperformed for the rest of the measures. Finally, data set size also plays a great role in performance, as Fast XML has the upper hand for small scale data sets while PfastreXML performs better, for large scale tasks.

In total, Subset LLDA manages to improve substantially over the state-of-the-art LLDA algorithms, while at the same time being competitive with two of the best performing extreme classification methods, under a variety of settings and evaluation measures.

## 5 Conclusions and Future Work

In this work we have presented an extension of LLDA, to account for large-scale and extreme classification tasks. Our algorithm Subset LLDA, proceeds in two stages, by first determining a set of relevant labels for a given test instance, and then constraining the CGS-LLDA algorithm to search only this label subspace for a solution. Extensive experiments on four small scale and four large scale data sets, with label sets sizes ranging from hundreds to hundreds of thousands, show a significant improvement over the best performing LLDA-based algorithms and competitive results with the state-of-the-art in extreme classification.

|  | FastXML | PfastreXML | PriorLDA | DepLDA | Subset LLDA |
|---|---|---|---|---|---|
| Bibtex | **0.586**▽ | 0.524 | 0.485 | 0.426 | 0.495 |
| Delicious | **0.357**▽ | 0.333 | 0.258 | 0.263 | 0.278 |
| Mediamill | 0.638 | **0.644**▽ | 0.547 | 0.536 | 0.622 |
| EUR-Lex | 0.319 | **0.411**▽ | 0.300 | 0.322 | 0.378 |
| BioASQ | 0.233 | 0.279 | - | - | **0.284** |
| Wiki10 | 0.105 | **0.184**▽ | - | - | 0.155 |
| Delicious-200k | **0.083** | 0.044 | - | - | 0.069 |
| Amazon | 0.153 | **0.321**▽ | - | - | 0.220 |
| Avg. Rank | 2.125 | 1.625 |  |  | 2.250 |

Table 8: Propensity Scored Precision@5 results for the LLDA-based and the extreme classification methods. A ▽ indicates a statistically significant difference between Subset LLDA and the best performing method, with a z-test and a significance level of 0.05. The - sign is used if the algorithm could not deliver predictions after 48 hours.

We would like to extend this work in several aspects: First, since our algorithm, as well as the rest of the algorithms considered in this work, can improve performance with more samples (or more trees in the case of Fast XML and PfastreXML), we would like to investigate in detail the time duration needed by each of the algorithms, to achieve a specific performance level. Additionally, our algorithm can be extended and improved in a number of ways. We have employed standard CGS for our LLDA model but we would like to use also two recently proposed faster LDA variants, WarpLDA [5] and LightLDA [28] that can scale to tasks with billions of parameters. Finally, instead of employing tf-idf to retrieve similar instances, we would like to also experiment with word2vec [13] or LDA.